\documentclass[10pt,twocolumn,letterpaper]{article}

\usepackage{cvpr}              

\usepackage{kotex}

\usepackage{adjustbox}
\usepackage{booktabs, makecell, tabularx}
\usepackage{siunitx}
\setcellgapes{3.5pt}
\usepackage{threeparttable}
\usepackage{algorithm} 
\usepackage[noend]{algpseudocode} 
\usepackage{footnote} 
\usepackage{float} 
\usepackage{wrapfig}

\usepackage{bbm}
\usepackage{cancel} 
\usepackage{xcolor} 

\usepackage{graphicx}
\usepackage{subcaption}

\usepackage{balance}

\definecolor{cvprblue}{rgb}{0.21,0.49,0.74}
\usepackage[pagebackref,breaklinks,colorlinks,allcolors=cvprblue]{hyperref}

\def\paperID{10434} 
\def\confName{CVPR}
\def\confYear{2026}

\title{
Post-Training and Test-Time Scaling of Generative Agent Behavior Models \\for Interactive Autonomous Driving
}

\author{
Hyunki Seong$^{1,2,\dagger}$\thanks{Work done during an internship at Qualcomm.}\quad Jeong-Kyun Lee$^{1\ddagger}$\quad Heesoo Myeong$^1$\quad Yongho Shin$^1$\\
Hyun-Mook Cho$^1$\quad Duck Hoon Kim$^1$\quad Pranav Desai$^{1\ddagger}$\quad Monu Surana$^{1\ddagger}$ \\ \\
$^1$Qualcomm \quad
$^2$Korea Advanced Institute of Science and Technology (KAIST)\\
{\tt\small $^\dagger$hynkis@kaist.ac.kr \quad $^\ddagger$\{ljeongky,pranavd,msurana\}@qti.qualcomm.com}
}

\begin{document}
\maketitle
\begin{abstract}
Learning interactive motion behaviors among multiple agents is a core challenge in autonomous driving. While imitation learning models generate realistic trajectories, they often inherit biases from datasets dominated by safe demonstrations, limiting robustness in safety-critical cases. Moreover, most studies rely on open-loop evaluation, overlooking compounding errors in closed-loop execution.
We address these limitations with two complementary strategies.
First, we propose Group Relative Behavior Optimization (GRBO), a reinforcement learning post-training method that fine-tunes pretrained behavior models via group relative advantage maximization with human regularization. Using only 10\% of the training dataset, GRBO improves safety performance by over 40\% while preserving behavioral realism.
Second, we introduce Warm-K, a warm-started Top-K sampling strategy that balances consistency and diversity in motion selection. Our Warm-K method-based test-time scaling enhances behavioral consistency and reactivity at test time without retraining, mitigating covariate shift and reducing performance discrepancies. Demo videos are available in the supplementary material.
\end{abstract}    
\section{Introduction}
\label{sec:intro}
Multi-agent motion generation for autonomous driving in urban environments is challenging due to complex dynamics and diverse inter-agent interactions involving vehicles, cyclists, and pedestrians. Recent advances address this by introducing simulation agent (Sim Agent) models \citep{suo2021trafficsim,zhong2022guided,montali2023waymo}, learning-based generative frameworks that predict and generate motion trajectories for multiple agents. These models capture rich spatial-temporal interactions while adhering to road semantics and driving behaviors.

Inspired by the success of large language models (LLMs), recent works \citep{seff2023motionlm,wu2024smart,zhang2025closed} have adopted GPT-style architectures and training paradigms for multi-agent motion prediction.
By representing motion trajectories as token sequences analogous to words in LLMs, these approaches employ decoder-only networks that generate future motions through the next-token prediction (NTP) paradigm \citep{chen2024next}, thereby enabling scalable sequence modeling with established language-modeling techniques.

Despite their success, these models are mainly trained through supervised imitation learning (IL), mimicking behaviors from human demonstrations. However, because such data predominantly consists of collision-free routine maneuvers over short horizons, the resulting models inherit biases that limit behavioral robustness, particularly in rare or safety-critical cases that are underrepresented in the dataset.
Furthermore, most existing methods rely on open-loop evaluation, generating single-shot predictions that fail to capture dynamic replanning in interactive environments. This overlooks compounding errors from behavioral inconsistencies in next-token sampling during closed-loop operation, leading to substantial performance gaps between open-loop evaluation and closed-loop deployment.

To overcome these limitations, we explore the Sim Agent models as world models capable of simulating diverse interactive driving situations. Leveraging their generative capabilities, the models perform self-simulation of motion rollouts with reward computation and trajectory evaluation. This facilitates \textit{self-policy improvement} through reinforcement learning (RL), guiding the policy model toward safer and more effective behaviors.
Moreover, with well-suited sampling strategies, generative policy models can produce diverse motion plans at test time, improving the possibility of yielding desirable solutions under specified criteria during closed-loop execution.

\begin{figure*}[t]
\centering
\includegraphics[width=0.99\textwidth]{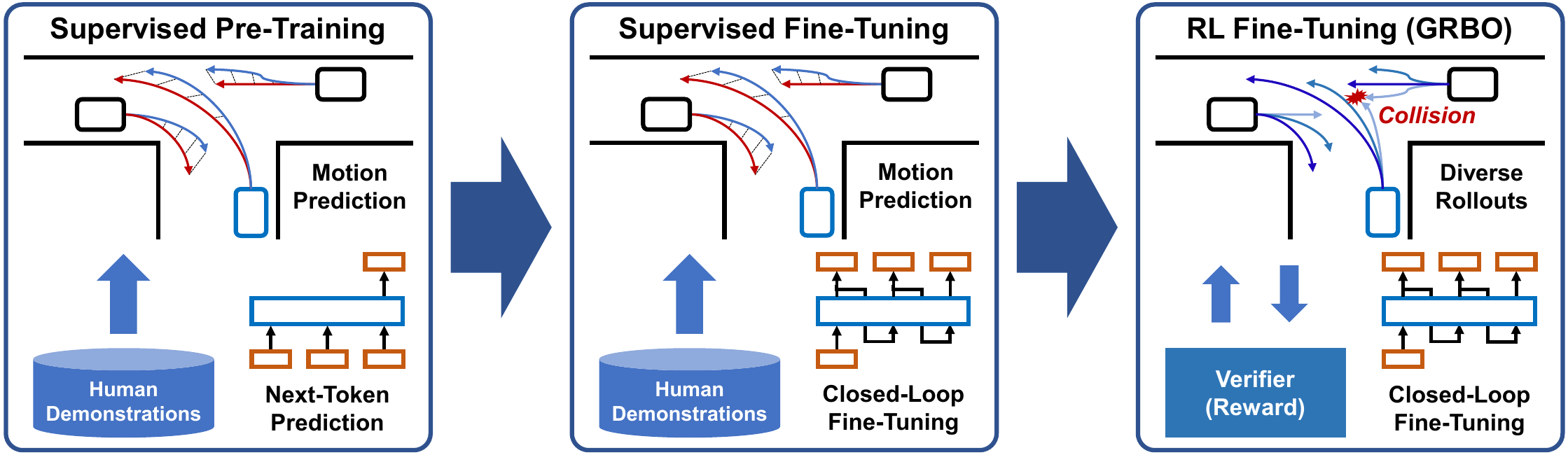}
\vspace{-8pt}
\caption{
We propose a novel RL-based post-training strategy that improves behavior planning performance while preserving the realistic features of pre-trained models. Our method is broadly applicable, including to supervised and other fine-tuned policies.
}
\label{fig:overview}
\vspace{-10pt}
\end{figure*}

In this paper, we introduce Group Relative Behavior Optimization (GRBO), a post-training framework for generative agent behavior models. GRBO enhances multi-agent motion planning by leveraging self-simulation and group-wise reward signals to refine policies beyond supervised imitation. Building on the group-relative policy optimization paradigm \citep{shao2024deepseekmath}, we develop an RL fine-tuning method for post-training interactive motion generation in multi-agent urban driving. Our approach substantially improves safety-critical performance in both common and the top 10\% high-risk urban driving scenarios, including long-tail cases, while preserving the model’s pretrained realism. Remarkably, GRBO achieves over a 40\% reduction in collision rate compared to supervised baselines, while requiring only 10\% of the original training data.

To further mitigate inconsistency in closed-loop environments, we introduce Warm-Started Top-K (Warm-K) sampling, a simple yet effective strategy to \textit{warm-start} next-token prediction. Leveraging prior motion selections, Warm-K sampling identifies the $K$ most likely motion tokens and selects the one best aligned with preceding plans. By combining warm-started and standard sampling, our method balances behavioral consistency and reactivity, achieving an average 43\% improvement in progress and 37\% reduction in acceleration during closed-loop execution without additional training.

Our contributions are as follows:
(1) We introduce GRBO, an RL-based post-training method for generative agent models in autonomous driving that achieves significant gains with only a fraction of the training data.
(2) We investigate the exploration–realism trade-off in post-training, showing that GRBO improves policy performance through exploration while preserving pre-trained human-likeness.
(3) We propose Warm-K, a warm-started token-sampling strategy that scales motion rollouts to balance consistency and reactivity, improving progress and efficiency in closed-loop execution at test time.
\section{Related Works}
\label{sec:related}

\noindent \textbf{Interactive Motion Generation.}
Classical interactive motion generation ranges from rasterized map-based prediction \citep{cui2019multimodal,xu2022bits} to joint motion forecasting conditioned on map and multi-agent context \citep{ngiam2021scene, nayakanti2022wayformer, shi2024mtr++}.
Recent Sim Agent approaches recast multi-agent behavior generation as autoregressive sequence modeling with Gaussian Mixture Models \citep{wang2023multiverse} or tokenized agent motions \citep{philion2023trajeglish,seff2023motionlm,wu2024smart}, using encoder-decoder transformers trained via supervised imitation learning.
While these IL-only models generate realistic rollouts through next-token prediction, they largely inherit safe-driving priors and short planning horizons, under-exploring rare yet safety-critical behaviors.
Self-play agents \citep{cornelisse2025building,cusumano2025robust} discover diverse behaviors through pure RL but often sacrifice human realism and demand costly closed-loop environment interaction.
Human-regularized RL \citep{cornelisse2024human} mitigates this by anchoring policies to human-like priors but requires a separate behavioral reference policy distinct from the RL model.
Our work follows the Sim Agent paradigm, autoregressive token policies over multi-agent context, but focuses on post-training that explicitly enhances safety while preserving human-likeness under interactive rollouts.

\noindent \textbf{Post-Training Strategies.}
Post-training has emerged in LLMs as a strategy for aligning and optimizing pretrained models \citep{ouyang2022training,achiam2023gpt}. Inspired by this success, recent agent behavior models have begun adapting similar post-training schemes to refine pretrained policies.
Supervised finetuning (SFT) with closed-loop rollouts \citep{zhang2025closed} stabilizes the realism of agent motions, though its performance remains limited by the quality and coverage of labeled data.
Several RL-based fine-tuning methods \citep{peng2024improving,ahmadirlftsim} use a classic RL method \citep{williams1992simple} to improve trajectory generation without human data. However, they often suffer from high-variance credit assignment inherent in the base RL algorithm.
To address this, we reformulate Group Relative Policy Optimization (GRPO) \citep{shao2024deepseekmath,liu2025understanding} for multi-agent motion generation. GRPO stabilizes policy updates via clipping \citep{schulman2017proximal} and replaces value-based advantage estimation with group-relative advantages that capture the relative superiority of rollouts within shared contexts. This design improves credit assignment and reduces variance without requiring a value network. Building on these benefits, our GRBO incorporates rewards to encourage human-aligned preferences such as safety while preserving pretrained realism through KL regularization. By optimizing pretrained models through group-wise rollout sampling, our method achieves self-policy improvement without relying on external simulation.

\noindent \textbf{Behavioral Consistency.}
Researchers have investigated behavioral and temporal consistency across diverse domains. In sequence modeling, LSTM and transformer-based methods \citep{lim2021temporal,tang2023semi} incorporate historical context to improve short-horizon coherence. In motion planning, RL-based adaptors \citep{fan2018hierarchical} bridge the gap between imitated and feasible trajectories, while action chunking with temporal ensembles \citep{zhao2023learning} promotes smooth long-horizon control at the cost of reactivity. More recently, bidirectional decoding \citep{liu2024bidirectional} samples multiple rollouts and selects temporally aligned plans via rollout-level scoring, though it requires sufficient sampling to include consistent candidates. Our Warm-K strategy introduces a token-level warm-start mechanism to select temporally aligned motion tokens during rollout generation, combining Warm-K and Top-K sampling to balance coherence and reactivity in closed-loop execution.
\section{Problem Definition}
\label{sec:problem_definition}
We first formulate a multi-agent behavior policy as a conditional distribution $\pi_{\phi}(\textbf{a}_{t}|\mathbf{s}_{\leq t}, \mathcal{M})$, where $\phi$ are learnable parameters, $\textbf{a}_t=[a_{1,t}, ..., a_{N,t}]$ the predicted motion tokens for $N$ agents at time $t$, $\mathbf{s}_{\leq t}$ the historical states, and $\mathcal{M}$ the scene context (e.g., road maps, traffic lights). The state $\mathbf{s}_t=[s_{1,t}, ..., s_{N,t}]$ represents the current configuration of all agents.
At each step, the policy samples the next agent motion token from the conditional distribution $\textbf{a}_t \sim \pi_\phi(\cdot \mid \mathbf{s}_{\leq t}, \mathcal{M})$. The sampled action is then applied to the environment or simulator $\mathbf{s}_{t+1} = f(\mathbf{s}_t, \textbf{a}_t)$, resulting in the next state $\mathbf{s}_{t+1}$.
Starting from $t=0$, we obtain a rollout of the agents' motions $\mathbf{s}_{0:T}$ of length $T$ through autoregressive sampling from the policy $\pi_{\phi}$.

\noindent \textbf{Generative Agent Models.}
We use SMART \citep{wu2024smart}, a decoder-only transformer framework for autonomous driving behavior generation. It encodes vectorized maps and agent trajectories into discrete action tokens and is trained with an NTP objective over spatio-temporal sequences.
The selected action $a_{i,t}$ is drawn from a vocabulary of motion tokens $\mathcal{V} = \{a^{c}_{i,t} | c = 1, ..., |\mathcal{V}|\}$, which induces a mapping between the continuous state space $\mathcal{S}$ and the discrete action-token space $\mathcal{A}$ via tokenization and detokenization.
This GPT-style approach captures motion distributions in real driving and generates diverse trajectories reflecting complex urban multi-agent interactions.
To train the behavior model, a batch of human demonstrations $\{\mathbf{s}^{GT}_{0:T}, \mathcal{M}\}$ is sampled from a dataset $\mathcal{D}$, where $\mathbf{s}^{GT}_{0:T}$ denotes the ground-truth (GT) state sequences of $N$ agents. For each agent, the corresponding GT motion-token action $a^{GT}_{i,t}$ is obtained from the given GT states.
The standard training objective is to learn $\pi_\phi$ in a supervised manner (e.g., IL) by minimizing the negative log-likelihood of the GT actions:
\begin{flalign}
    \label{eq:loss_ntp}
    \mathcal{L}_{NTP}(\phi) &= -\frac{1}{N} \!\sum^{N}_{i=1} \frac{1}{T} \!\sum^{T-1}_{t=0} \text{log} \pi_{\phi} (a^{GT}_{i,t} | s_{i,\leq t}, \mathcal{M}).
\end{flalign}

In this formulation, multi-agent motion generation is cast as a sequential prediction problem: given past trajectory history and scene context, the model autoregressively predicts future motions of all agents in a manner that is consistent and interactive over time.

\noindent \textbf{Reinforcement Learning.}
To enhance scalability and robustness, we extend the multi-agent behavior modeling task into a contextual Markov Decision Process (MDP), enabling RL-based fine-tuning. The goal of the RL problem is to improve the policy to generate safer motion trajectories for multiple interacting agents, while preserving the original socially consistent, scene-aware behavior.
The MDP is defined by the tuple $(\mathcal{S}, \! \mathcal{A}, \! f, \! \pi_{\phi}, \! R, \! \mathcal{X}_{\mathcal{M}})$, where $\mathcal{S}$ denotes the joint state space of agents, $\mathcal{A}$ is the joint action space, and $f:\! \mathcal{S} \! \times \! \mathcal{A} \! \rightarrow \! \mathcal{S}$ defines the transition dynamics. $R$ is a reward function, and $\mathcal{X}_{\mathcal{M}}$ denotes the map context space.
At each step, the policy $\pi_{\phi}$ autoregressively predicts the next joint action conditioned on the past states $\mathbf{s}_{\leq t}$ and the scene context $\mathcal{M}$. The initial past states and scene context $\{\mathbf{s}^{\mathrm{GT}}_{\leq 0}, \mathcal{M}\}$ are drawn from the dataset $\mathcal{D}$ and used to initialize the autoregressive rollout. A full trajectory $\mathbf{s}_{0:T}$ is then generated by rolling out the policy under the transition dynamics.
\section{Methodologies}
\label{sec:method}
\subsection{Group Relative Behavior Optimization}

\textbf{RL-based Post-Training.}
Fig. \ref{fig:overview} and Algorithm \ref{algo:grbo} summarize the overall training process. We extend the supervised agent behavior model with RL to refine closed-loop motion generation. Unlike IL, which directly mimics human trajectories, GRBO leverages the generative capacity of the SimAgent model for self-simulation: the policy autoregressively unrolls multiple inter-agent trajectories per traffic scenario and evaluates them with a reward model. These rollouts encompass both nominal driving and rare, safety-critical interactions that are difficult to obtain from demonstrations, enabling explorative policy improvement without online interaction. GRBO further performs group-wise comparisons among rollouts from the same inputs, facilitating relative advantage estimation and guiding the policy toward safer and more optimal urban driving behaviors.

\noindent \textbf{Objective Function.}
The learning objective follows the clipped policy optimization framework, augmented with group-relative advantages inspired by the GRPO algorithm \citep{shao2024deepseekmath}. Each candidate rollout within an urban traffic scenario is scored relative to other rollouts in the same group, allowing the model to emphasize relative improvements over absolute, potentially noisy scores. The full GRBO objective (Eq. \ref{eq:objective}) balances three terms: (i) RL updates guided by relative advantages, (ii) a clipping mechanism for stability, and (iii) KL-regularization to anchor the policy to the imitation-learned behavior and preserve human-likeness:

\begin{flalign}
    \label{eq:objective}
    \mathcal{J}&_{GRBO}(\phi) = \mathbb{E}_{(\textbf{s}_{\leq 0}, \mathcal{M}) \sim D, {\{\textbf{s}^j_{0:T}\}^{G}_{j=1}} \sim \pi_{\phi_{old}}(\cdot|\textbf{s}_{\leq 0}, \mathcal{M})}\\ \nonumber
    &\frac{1}{G} \!\sum^{G}_{j=1} \frac{1}{N} \!\sum^{N}_{i=1} \frac{1}{T} \!\sum^{T-1}_{t=0} \!\Bigl\{\text{min} \Big[r^j_{i,t}(\phi) \!\hat{A}^j_{i,t},\\ \nonumber
    &\text{clip} \Big(r^j_{i,t}(\phi), 1\!-\!\epsilon_{l}, 1\!+\!\epsilon_{h} \Big) \!\hat{A}^j_{i,t} \Big]\! -\! \beta \mathbb{D}_{\text{KL}}(\pi_{\phi} | \pi_{ref}) \Bigl\},
\end{flalign}
where
\renewcommand\CancelColor{\color{red}} 
\vspace{-15pt}
\begin{flalign}
    \label{eq:policy_ratio}
    &r^j_{i,t}(\phi) = \frac{\pi_{\phi}(a^j_{i,t}|s^j_{i,\leq t}, \mathcal{M})}{\pi_{\phi_{old}}(a^j_{i,t}|s^j_{i,\leq t}, \mathcal{M})},\\
    \label{eq:advantage}
    &\hat{A}^j_{i,t} = \frac{R^j_{i} - \text{mean}(\{R^j_{i}\}^{G}_{j=1})}{\CancelColor{\xcancel{{\text{std}(\{R^j_{i}\}^{G}_{j=1})}}}}.
\end{flalign}

\noindent {\textbf{Scenario-level Difficulty Bias.}}
We remove the standard deviation term from the relative advantage calculation (Eq. \ref{eq:advantage}) for two reasons.
(i) Similar to the language domain \citep{liu2025understanding}, multi-agent interactive motion generation exhibits scenario-level difficulty bias, where nominal or safety-critical conditions produce low reward variance in overly easy or difficult scenarios, unintentionally assigning higher advantage weights and introducing bias into policy optimization.
(ii) We note that groups with higher reward variance are more informative, which typically contain a few samples with exceptionally high or low rewards. Such diversity enables the policy to encourage previously unexplored desirable actions while discouraging rare but critical failures, thereby leading to substantial performance improvements.

\noindent \textbf{Human Regularization.}
The Kullback-Leibler (KL) term in Eq. \ref{eq:objective} ensures that the post-trained model does not drift excessively from the pre-trained human-like distribution. This human regularization is critical for maintaining realism, as purely optimizing for safety can lead to overly conservative or unnatural behaviors. By penalizing divergence from the reference policy, GRBO enforces a trade-off between performance improvement and behavioral fidelity.
Unlike the original KL penalty defined between two probability distributions in prior work \citep{cornelisse2024human}, we approximate the divergence using the following unbiased estimator \citep{kl_approx}:
\vspace{-5pt}
\begin{flalign}
    \label{eq:kl}
    &\mathbb{D}_{\text{KL}}(\pi_{\phi} | \pi_{ref}) \\ \nonumber
    &= \frac{\pi_{\text{ref}}(a^j_{i,t}|s^j_{i,\leq t}, \mathcal{M})}{\pi_{\phi}(a^j_{i,t}|s^j_{i,\leq t}, \mathcal{M})} - \text{log} \frac{\pi_{\text{ref}}(a^j_{i,t}|s^j_{i,\leq t}, \mathcal{M})}{\pi_{\phi}(a^j_{i,t}|s^j_{i,\leq t}, \mathcal{M})} - 1,
\end{flalign}
which is computationally efficient for autoregressive next-token prediction methods.

\noindent \textbf{Reward Function.}
The reward function is defined as
\vspace{-2pt}
\begin{flalign}
    \label{eq:reward}
    R_i^j = -\,\mathbb{I}\!\left[\exists\, t \in [1, T] : \textit{Colli}_{i,t}^j = 1\right],
\end{flalign}
 where $\textit{Colli}_{i,t}^j$ denotes a Boolean collision indicator for agent $i$ at time step $t$. This formulation penalizes any trajectory that experiences at least one collision during the rollout, directly encouraging safety-preserving behaviors.
 This binary reward provides a strong signal for safety-critical improvement, as collision avoidance remains the primary objective.
 The group-relative advantage normalization further ensures fair reward comparison within sampled rollouts, stabilizing gradient updates.

\subsection{Rollout Sampling Strategies}
\noindent \textbf{Top-K Sampling for Post-Training.}
During RL fine-tuning, we employ Top-K random sampling \citep{fan2018hierarchical} to generate diverse candidate rollouts, retaining only the most probable motion tokens at each step. This reduces variance from unlikely outliers and focuses training on plausible yet diverse behaviors, balancing exploration and tractability.

\begin{algorithm}[!t]
\caption{Group Relative Behavior Optimization}
\label{algo:grbo}
\small
\textbf{Input: } Pre-trained policy $\pi_{\phi_{\text{init}}}$, dataset $D$ \\
\textbf{Output: } \text{Post-trained behavior model } $\pi_{\phi}$

\begin{algorithmic}[1] 
\State Behavior model $ \pi_{\phi} \leftarrow \pi_{\phi_\text{init}} $
\State Reference model $ \pi_{\text{ref}} \leftarrow \pi_{\phi_\text{init}} $ \Comment{\textcolor{blue}{Human regularization.}}

\For{each iteration} \Comment{\textcolor{blue}{Closed-loop reinforcement learning.}}
    \State Sample a batch of data $\{\textbf{s}_{\leq 0}, \mathcal{M}\}$ from $D$
    \State Update the old behavior model $ \pi_{\phi_{\text{old}}} \leftarrow \pi_{\phi} $
    \For{$t = 0, ..., T-1$} \Comment{\textcolor{blue}{$T$ steps autoregressive rollout.}}
            \State Sample $G$ next motion tokens for $N$ agents $ \{\textbf{a}^{j}_{t}\}^{G}_{j=1} $.
            \State Get a group of next rollout states $ \{\textbf{s}^{j}_{t}\}^{G}_{j=1} $.
    \EndFor
    \State Compute rewards $R^j_{i}$ for each sampled rollout $s^j_{i,0:T}$.
    \State Compute $A^j_{i,t}$ for the $t$-th motion token of $s^j_{i,0:T}$ through group relative advantage estimation (Eq. \ref{eq:advantage}).
    \State Calculate $\mathbb{D}_{\text{KL}}(\pi_{\phi} | \pi_{ref})$ between the current and reference behavior models via Eq. \ref{eq:kl}.
    \State Update $\phi$ by minimizing $\mathcal{J}_{\text{GRBO}}(\phi)$ (Eq. \ref{eq:objective})
\EndFor
\end{algorithmic}
\end{algorithm}

\noindent \textbf{Warm-K Sampling at Test Time.}
\begin{figure*}[t]
\centering
\includegraphics[width=0.99\textwidth]{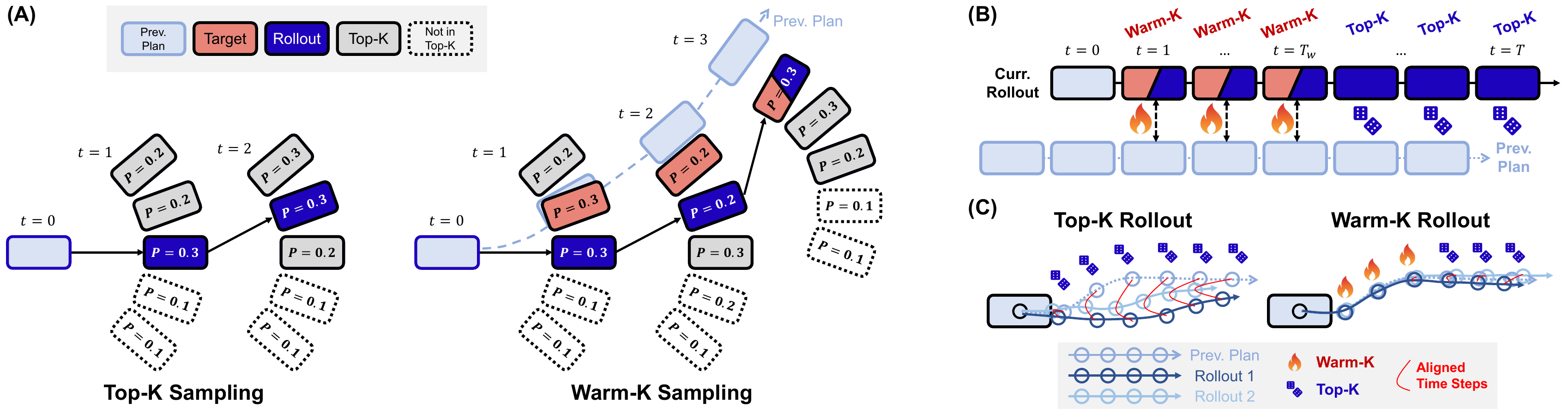}
\vspace{-5pt}
\caption{
\textbf{(A):} Top-K vs. Warm-K sampling.
\textbf{(B):} Warm-K warm-starts next-token prediction using prior plans in early steps, then switches to Top-K for reactivity and diversity.
\textbf{(C):} Top-K yields diverse but inconsistent rollouts, while Warm-K balances consistency and diversity.
}
\label{fig:warm-k}
\vspace{-12pt}
\end{figure*}
To address behavioral inconsistencies in next-token sampling, we propose Warm-Started Top-K (Warm-K), a test-time sampling method that leverages historical motion context to warm-start token selection, inspired by warm-start optimization \citep{yildirim2002warm}. As illustrated in Fig. \ref{fig:warm-k}, instead of cold-start sampling, candidate tokens are drawn from the top-k set and biased toward those consistent with prior motion choices, thereby improving behavioral consistency.
Our method builds on the $\text{top}^{K}$ operator formulation in \citep{zhang2025closed}, with a key modification: when identifying the target action token, we use next states ($\mathbf{s}_{t+1}$) from \textit{the temporally aligned previous plan} to generate coherent rollouts, rather than ground-truth next states ($\mathbf{s}^{\mathrm{GT}}_{t+1}$), which are unavailable at test time.
During inference, Warm-K is applied in the early phase for $T_w$ steps to maintain consistency, then transitions to standard Top-K to enhance reactivity and diversity, achieving a balanced trade-off between consistency and responsiveness in closed-loop execution.

\subsection{Closed-Loop Planning with Test-Time Scaling}
We formalize test-time scaling as the process of adjusting rollout diversity and selection criteria during inference to mitigate covariate shift. Standard autoregressive rollouts tend to accumulate compounding errors; thus, we explore scaling in two directions: (1) \textit{Consistency Scaling}: using the Warm-K-based hybrid strategy for warm-started sampling toward coherent trajectories; and (2) \textit{Reactivity Scaling}: retaining standard Top-K candidates to allow diverse and reactive maneuvers.
By adaptively mixing these sampling modes, our method scales the policy at test time, enhancing closed-loop robustness without additional training.
After applying the two rollout scaling methods, we perform Best-of-N selection \citep{gui2024bonbon} using the following score function as the criterion:
\vspace{-5pt}
\begin{flalign}
    \label{eq:score}
    S^j_{ego} = - \frac{1}{T} \sum^{T-1}_{t=0} \{ w_{c} \textit{Colli}^j_{t} + w_{a} \textit{Accel}^j_{t} \},
\end{flalign}
where $\textit{Accel}^j_{t}$ denotes acceleration, and $w_{c}$ and $w_{a}$ are the corresponding weights. We include $\textit{Accel}^j_{t}$ to account for driving efficiency, which was not part of the reward function, enabling investigation of \textit{whether test-time scaling can handle factors absent during RL post-training}.
In the closed-loop setting, we use the ego’s single-agent score but can extend it to global multi-agent scores, an avenue for further improvement.
At each step, only the current action from the selected rollout is executed, while the remainder is discarded as the policy replans from updated states following the receding horizon planning (RHP) scheme \citep{mayne1988receding}.

\section{Experiments}
\label{sec:experiments}

\begin{table*}[t] 
\caption{Quantitative Evaluation (2\% Validation Split)}
\vspace{-0.5em}
\centering
\begin{adjustbox}{width=0.99\textwidth, pagecenter}
\renewcommand{\arraystretch}{1.2}
\begin{tabular}{c*{1}{S}|c*{5}{S}}
\toprule
Method
    & {\makecell{Strategy}}
    & {\makecell{Collision Rate}}
        & {\makecell{Realism Meta M.}}
            & {\makecell{Interactive M.}}
                & {\makecell{Map-based M.}}
                    & {\makecell{Kinematic M.}} \\
\cmidrule(lr){1-1}
\cmidrule(lr){2-2}
\cmidrule(lr){3-3}
\cmidrule(lr){4-7}
SMART   &
IL  &
\text{0.0482 \quad \ \ (-)\quad \ } & \text{0.7655 \quad \ \ (-)\quad \ } & \text{0.8075 \quad \ \ (-)\quad \ } & \text{0.8707 \quad \ \ (-)\quad \ } & \text{0.4864 \quad \ \ (-)\quad \ } \\
CAT-K   &
IL \& SFT &
\text{0.0438 (\ \ -9.24\%)} & \text{0.7668 (+0.17\%)} & \text{0.8087 (+0.15\%)} & \text{0.8724 (+0.20\%)} & \text{0.4873 (+0.19\%)} \\
REINFORCE   &
IL \& RLFT &
\text{0.0354 (\ -26.61\%)} & \text{0.7547 (-1.40\%)} & \text{0.8035 (-0.50\%)} & \text{0.8613 (-1.08\%)} & \text{0.4586 (-5.71\%)} \\
\cmidrule(lr){1-1}
\cmidrule(lr){2-2}
\cmidrule(lr){3-3}
\cmidrule(lr){4-7}
GRBO   &
IL \& RLFT &
\text{0.0270 (-44.08\%)} & \text{0.7634 (-0.27\%)} & \text{0.8063 (+0.15\%)} & \text{0.8698 (-0.10\%)} & \text{0.4806 (-1.20\%)} \\
CAT-K \& GRBO   &
IL \& SFT \& RLFT &
\text{0.0290 (-39.91\%)} & \text{0.7637 (-0.23\%)} & \text{0.8073 (-0.03\%)} & \text{0.8686 (-0.24\%)} & \text{0.4821 (-0.88\%)} \\
\bottomrule

\end{tabular}
\end{adjustbox}
    \label{table:eval-valid-split}
\vspace{-1.0em}
\end{table*}

\begin{table*}[!t] 
\caption{Quantitative Evaluation (Overall-3000)}
\vspace{-0.5em}
\centering
\begin{adjustbox}{width=0.99\textwidth, pagecenter}
\renewcommand{\arraystretch}{1.0}
\begin{tabular}{c*{1}{S}|c*{5}{S}}
\toprule

Method
    & {\makecell{Strategy}}
    & {\makecell{Collision Rate}}
        & {\makecell{Realism Meta M.}}
            & {\makecell{Interactive M.}}
                & {\makecell{Map-based M.}}
                    & {\makecell{Kinematic M.}} \\
\cmidrule(lr){1-1}
\cmidrule(lr){2-2}
\cmidrule(lr){3-3}
\cmidrule(lr){4-7}
SMART   &
IL  &
\text{0.0413 \quad \ \ (-)\quad \ } & \text{0.7694 \quad \ \ (-)\quad \ } & \text{0.8102 \quad \ \ (-)\quad \ } & \text{0.8758 \quad \ \ (-)\quad \ } & \text{0.4913 \quad \ \ (-)\quad \ } \\
CAT-K   &
IL \& SFT &
\text{0.0390 (\ -5.63\%)} & \text{0.7710 (+0.21\%)} & \text{0.8121 (+0.24\%)} & \text{0.8769 (+0.12\%)} & \text{0.4932 (+0.38\%)} \\
\cmidrule(lr){1-1}
\cmidrule(lr){2-2}
\cmidrule(lr){3-3}
\cmidrule(lr){4-7}
GRBO-E3   &
IL \& RLFT &
\text{0.0255 (-38.24\%)} & \text{0.7676 (-0.23\%)} & \text{0.8119 (+0.22\%)} & \text{0.8723 (-0.39\%)} & \text{0.4847 (-1.35\%)} \\
GRBO  &
IL \& RLFT &
\text{0.0230 (-44.33\%)} & \text{0.7673 (-0.27\%)} & \text{0.8108 (+0.07\%)} & \text{0.8726 (-0.37\%)} & \text{0.4852 (-1.24\%)} \\
\bottomrule

\end{tabular}
\end{adjustbox}
    \label{table:eval-overall}
\vspace{-1.0em}
\end{table*}

\begin{table*}[!t] 
\caption{Quantitative Evaluation (Top-10\% Safety-Critical)}
\vspace{-0.5em}
\centering
\begin{adjustbox}{width=0.99\textwidth, pagecenter}
\renewcommand{\arraystretch}{1.0}
\begin{tabular}{c*{1}{S}|c*{5}{S}}
\toprule
Method
    & {\makecell{Strategy}}
    & {\makecell{Collision Rate}}
        & {\makecell{Realism Meta M.}}
            & {\makecell{Interactive M.}}
                & {\makecell{Map-based M.}}
                    & {\makecell{Kinematic M.}} \\
\cmidrule(lr){1-1}
\cmidrule(lr){2-2}
\cmidrule(lr){3-3}
\cmidrule(lr){4-7}
SMART   &
IL  &
\text{0.2487 \quad \ \ \ (-)\quad \ } & \text{0.7396 \quad \ \ (-)\quad \ } & \text{0.7721 \quad \ \ (-)\quad \ } & \text{0.8530 \quad \ \ (-)\quad \ } & \text{0.4680 \quad \ \ (-)\quad \ } \\
CAT-K   &
IL \& SFT &
\text{0.2422 (\ \ -2.61\%)} & \text{0.7433 (+0.51\%)} & \text{0.7760 (+0.51\%)} & \text{0.8568 (+0.45\%)} & \text{0.4711 (+0.68\%)} \\
\cmidrule(lr){1-1}
\cmidrule(lr){2-2}
\cmidrule(lr){3-3}
\cmidrule(lr){4-7}
GRBO-E3  &
IL \& RLFT &
\text{0.1826 (-26.57\%)} & \text{0.7342 (-0.73\%)} & \text{0.7721 (-0.01\%)} & \text{0.8454 (-0.88\%)} & \text{0.4541 (-2.97\%)} \\
GRBO &
IL \& RLFT &
\text{0.1712 (-31.17\%)} & \text{0.7299 (-1.31\%)} & \text{0.7646 (-0.97\%)} & \text{0.8438 (-1.08\%)} & \text{0.4524 (-3.32\%)} \\
\bottomrule

\end{tabular}
\end{adjustbox}
    \label{table:eval-safety}
\end{table*}

\begin{figure*}[!t]
\centering
\includegraphics[width=0.99\textwidth]{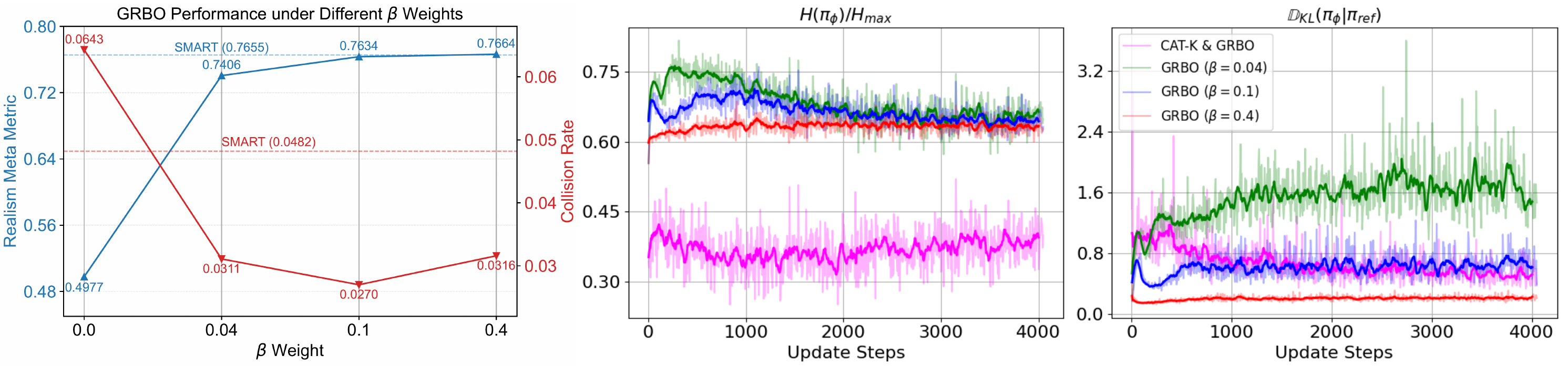}
\caption{
\textbf{Left: } Performance comparison between different KL weights $\beta$. \textbf{Middle: } Normalized entropy curves during post-training. \textbf{Right: } The KL divergence during post-training. 
}
\label{fig:plot_tradeoff}
\vspace{-15pt}
\end{figure*}

\subsection{Experimental Setup}
We train and evaluate our approach on the Waymo Open Motion Dataset (WOMD) \citep{ettinger2021large}, a large-scale urban driving dataset. Following SMART, we adopt its network architecture as the policy model and pre-train it with supervised learning for 32 epochs on the full dataset. For post-training, we fine-tune the model using RL with only 10\% of the original data for 10 epochs, highlighting both training efficiency and the exploratory advantages of RL. We apply gradient accumulation during the group-sampling-based RL stage to ensure equivalent computational conditions (total batch size of 80 on 8 × A100 80 GB GPUs) across the baseline pre-training and post-training methods. Closed-loop evaluations are conducted in Waymax \citep{gulino2023waymax}, an external simulator that provides a multi-agent interactive evaluation environment.

\noindent \textbf{Baseline Approaches.}
We evaluate our method against several baselines. SMART \citep{wu2024smart} serves as the supervised IL baseline, and CAT-K \citep{zhang2025closed} extends it with supervised fine-tuning (SFT) after IL pre-training. To incorporate RL, REINFORCE \citep{peng2024improving} is used as an RL fine-tuning method following IL pre-training. Our proposed GRBO applies group-relative optimization after IL pre-training to further refine policy performance. We also assess GRBO-E3, trained for three epochs, to analyze RL post-training efficiency. Finally, we examine a hybrid CAT-K \& GRBO approach that combines supervised fine-tuning with CAT-K rollouts and GRBO-based RL post-training.

\noindent \textbf{Metrics.}
We adopt complementary metrics to evaluate both open-loop prediction and closed-loop execution. In the open-loop setting, we report the \textit{Collision Rate}, measuring the frequency of collisions across rollouts, and the \textit{Realism Meta Metric}, a composite score from the Waymo Open Sim Agents Challenge (WOSAC) \citep{montali2023waymo} that quantifies the human-likeness of generated trajectories. We also include \textit{Interactive}, \textit{Map-based}, and \textit{Kinematic metrics} from the same benchmark, assessing social compliance, map adherence, and kinematic similarity, respectively. In the closed-loop setting, we evaluate dynamic driving performance using \textit{Progress}, defined as the terminal progress ratio to goal, and \textit{Acceleration}, the mean absolute acceleration over the rollout, reflecting behavioral efficiency and driving comfort.

\noindent \textbf{Evaluation Conditions.}
For open-loop evaluation, we use three validation sets: \textit{2\% Validation Split} for consistency with the baseline study \citep{zhang2025closed}, \textit{Overall-3000} for large-scale evaluation across 3,000 scenes, and \textit{Top-10\% Safety-Critical} for urban scenarios with the highest 10\% collision likelihood under the SMART baseline.
For closed-loop evaluation, agents are deployed in Waymax-based interactive environments for up to 80 steps (8 seconds), matching the open-loop horizon. Each agent’s state evolves through the transition function $f$ according to the action selected by the RHP scheme. To promote consistent motion generation, the behavior models are trained with an additional goal input inserted before the final token-selection layer.

\begin{figure*}[t]
\centering
\includegraphics[width=0.95\textwidth]{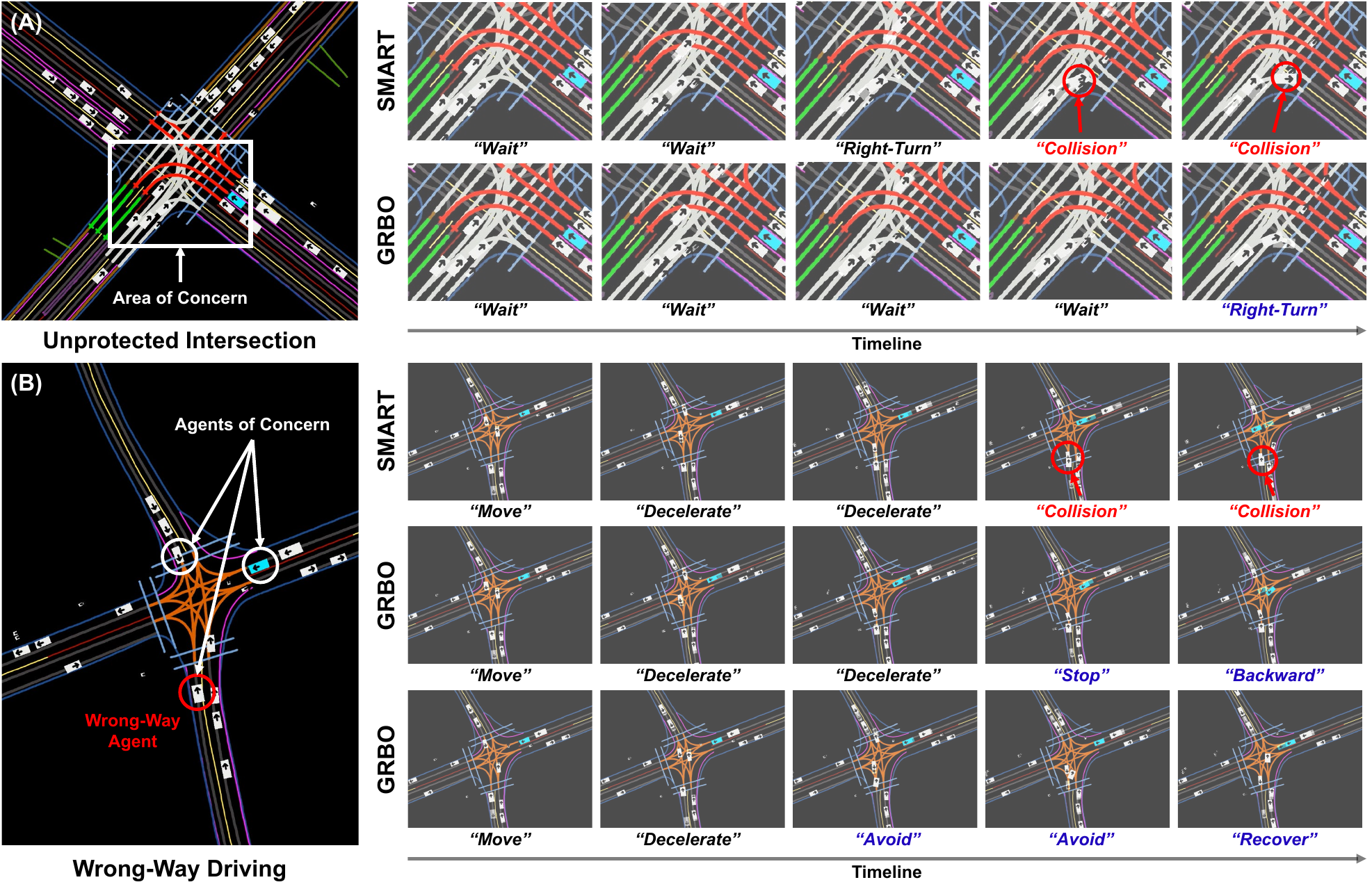}
\vspace{-5pt}
\caption{
Simulation results in long-tail safety-critical cases. \textbf{(A): } A right-turn scenario in a congested traffic environment where interactions between vehicles and cyclists are highly active. \textbf{(B): } A long-tail intersection scenario in which an agent drives the wrong way.
}
\label{fig:demo-long-tail}
\vspace{-10pt}
\end{figure*}

\begin{figure*}[!t]
\centering
\includegraphics[width=0.95\textwidth]{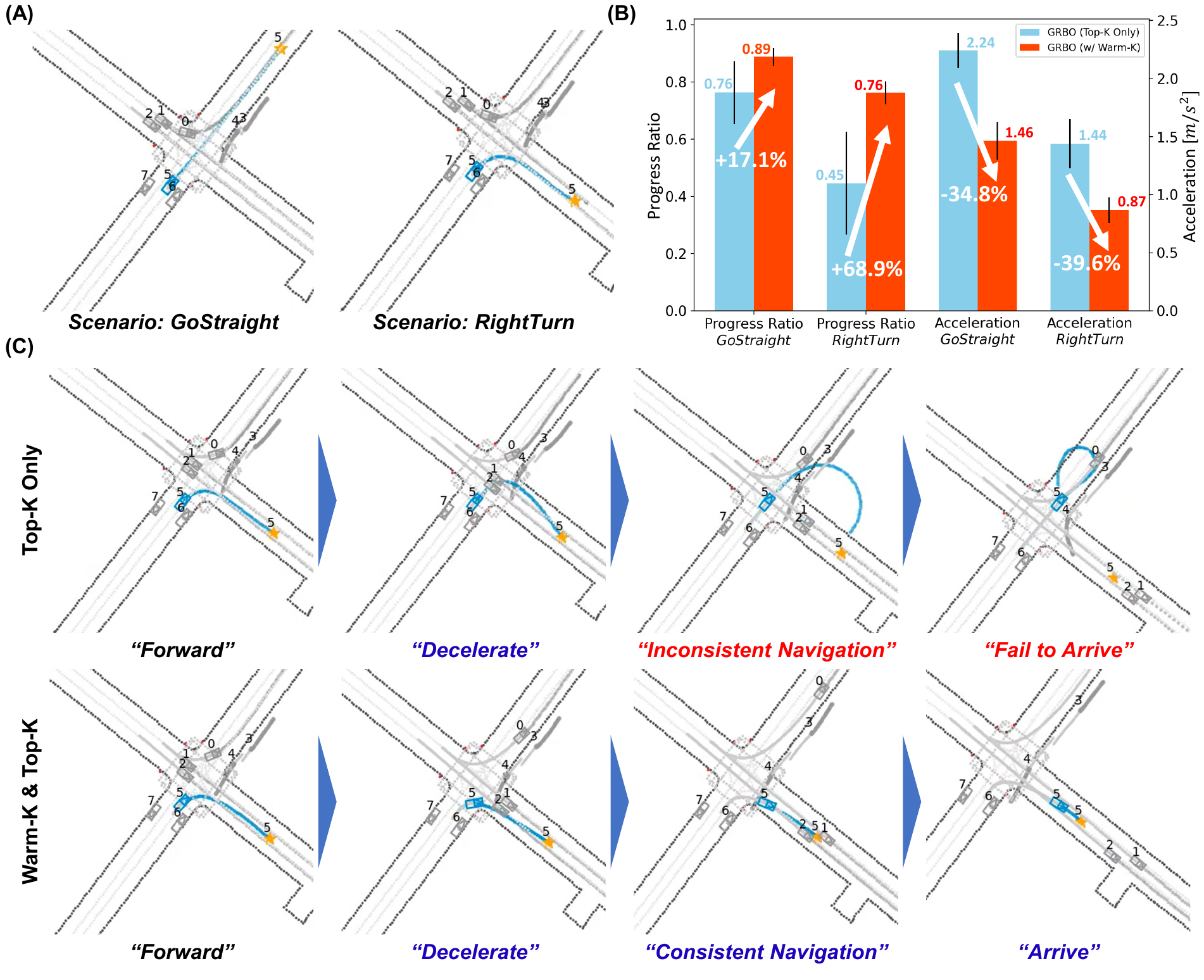}
\vspace{-5pt}
\caption{
Open-loop (A) and closed-loop (B, C) results in intersections. Comparing Top-K and Warm-K hybrid sampling shows that the hybrid method improves behavioral consistency while enabling timely, collision-free navigation in both straight and right-turn scenarios.
}
\label{fig:closed-loop}
\vspace{-12pt}
\end{figure*}

\subsection{Open-Loop Performance Comparison}
\noindent \textbf{Overall Cases.}
When evaluated on both the 2\% validation split (Table \ref{table:eval-valid-split}) and the larger Overall-3,000 benchmark (Table \ref{table:eval-overall}), GRBO consistently demonstrates substantial safety improvements. In the validation split, collision rates dropped by over 44\% compared to the SMART baseline, and similar gains were reproduced at scale in the larger evaluation set, where GRBO lowered collisions by 38-44\%. Importantly, these safety improvements are achieved while maintaining nearly identical realism scores, indicating that the model’s human-likeness is preserved despite RL updates. Notably, even with only a few epochs of RL post-training (e.g., GRBO-E3), the collision rate is reduced by up to 38.24\%. This shows the robustness and generalizability of GRBO, whose safety benefits extend beyond small validation subsets to diverse urban scenarios, outperforming both purely supervised and RL fine-tuning baselines.

\noindent \textbf{Top-10\% Safety-Critical Cases.}
The benefits of GRBO become even more pronounced in rare but high-risk scenarios (Table \ref{table:eval-safety}). In these safety-critical cases, where the baseline SMART and SFT models suffer frequent failures, GRBO achieves over a 30\% reduction in collision rate using only 10\% of the original training data, demonstrating that relative, within-group updates offer a decisive advantage in guiding the model toward safe resolutions of complex interactions. Importantly, this improvement comes without a significant loss in realism, underscoring the effectiveness of group-relative optimization in long-tail settings.
Although the kinematic similarity metric shows noticeable variation as RL optimization alters agent maneuvers, the realism change remains around 1\%, indicating that our method preserves human-like motion generation while substantially improving safety-critical performance from a kinematic behavior perspective.

\subsection{Analysis in the Open-Loop Evaluation}
\noindent \textbf{Exploration vs. Human-likeness Trade-Off.} 
Fig. \ref{fig:plot_tradeoff} (left) shows that smaller KL weights ($\beta$) allow greater deviation from the reference policy, expanding exploration and reducing collisions but risking degradation in human-likeness. In contrast, larger $\beta$ constrains updates, preserving realism but limiting safety gains. The normalized entropy curves in Fig. \ref{fig:plot_tradeoff} (middle) illustrate this trade-off: normalized entropy serves as a proxy for exploration capacity, reflecting action diversity. RLFT initially increased token-level entropy, allowing the policy to “open up” and discover collision-averse behaviors, before gradually decreasing as GRBO converges toward safer modes. Excessive entropy reduction, however, made the policy overly deterministic and less adaptive. This explains why supervised fine-tuning (SFT), despite modest gains, sharply reduced entropy and exploration. Even when followed by RL, SFT-trained models recovered only partially (less than 50\% of the maximum entropy), yielding limited improvements (e.g., CAT-K \& GRBO reduced collisions by 33.8\% from CAT-K, whereas GRBO alone achieved over 44\%; see Table \ref{table:eval-valid-split}). In contrast, RLFT maintained sufficient exploration and achieved substantial performance gains, with its KL curve showing minimal catastrophic forgetting, consistent with recent findings from comparisons between SFT and RL-based methods \citep{shenfeld2025rlrazor}. Finally, Fig. \ref{fig:plot_tradeoff} (right) shows that KL divergence rised during exploration and stabilizes under the KL penalty, confirming effective human-likeness regularization. These results suggest that scalable and robust post-training for agent behavior modeling requires either RL-centric fine-tuning or a balanced combination of SFT and RLFT. Accordingly, we adopt a moderate $\beta$ that promotes early exploration while anchoring the policy to realistic behaviors, achieving the best safety–realism trade-off observed in Tables \ref{table:eval-valid-split}–\ref{table:eval-safety}.

\noindent \textbf{Long-Tail Safety-Critical Situations.}
Qualitative analyses in Fig. \ref{fig:demo-long-tail} illustrate cases where exploration yields substantial policy improvement in long-tail, safety-critical conditions.
In a congested right-turn scene with active vehicle–cyclist interactions involving over 35 agents (Fig. \ref{fig:demo-long-tail}A), the supervised baseline (SMART) failed to generate safe trajectories, resulting in severe collisions. In contrast, GRBO produced human-like conservative maneuvers before executing the unprotected right turn, successfully avoiding collisions even under complex multi-agent interactions.
We further analyzed a long-tail urban case involving a wrong-way driving agent (Fig. \ref{fig:demo-long-tail}B), constructed from ill-labeled data where the agent’s initial heading was opposite to the road direction.
While the IL baseline reacted late and triggered collision chains, GRBO exhibited anticipatory and adaptive behaviors, including deceleration, brief backward motion, evasive lateral movement, and timely lane re-entry, which were maneuvers rarely observed in human driving data.
These results show that group-relative updates enable the policy to distinguish near-misses from safe resolutions within the same context, fostering behaviors that are both feasible and interaction-aware. Moreover, GRBO discovered novel recovery maneuvers in rare, safety-critical situations, aligning with the Top-10\% safety-critical results, where it achieved the largest collision-rate reduction while preserving realism.

\subsection{Closed-Loop Motion Planning Performance}
\noindent \textbf{Performance Discrepancies.}
We further analyzed the performance gaps between open- and closed-loop evaluations in two unprotected intersection cases: GoStraight and RightTurn.
In the open-loop setting (Fig. \ref{fig:closed-loop}A), the model consistently generated goal-reaching trajectories with a 100\% progress ratio through autoregressive motion selection.
In contrast, under closed-loop evaluation (Fig. \ref{fig:closed-loop}B), progress dropped to 76\% in the go-straight and 45\% in the right-turn cases.
These gaps highlight the compounding effect of distributional shift: while open-loop operation assumes ideal autoregression, closed-loop execution requires continual replanning, where small drifts accumulate into large behavioral deviations that hinder goal attainment.

\noindent \textbf{Behavioral Consistency and Efficiency.}
Fig. \ref{fig:closed-loop}B shows that the hybrid Warm-K \& Top-K sampling with test-time scaling enhances closed-loop motion generation by improving behavioral consistency while maintaining reactivity.
Across both scenarios, our method achieved over 17\% higher progress ratios than pure Top-K sampling, indicating more successful and timely maneuver completion. Meanwhile, the average acceleration decreased by up to 35\%, reflecting smoother and more efficient driving. The standard deviations of performance also decreased, suggesting that the warm-started strategy improved the stability and reliability of motion planning.
As shown in Fig. \ref{fig:closed-loop}C, compared with Top-K sampling, which often produces inconsistent rollouts that hinder navigation, our hybrid Warm-K strategy ensures consistent behavior, reduces unnecessary acceleration fluctuations, and reactively avoids collisions by selecting the best-performing motion plans among Warm-K and Top-K rollouts. These results demonstrate that the test-time scaling enhances both consistency and efficiency, yielding agents that are more reliable in closed-loop execution.
\section{Conclusion}
\label{sec:conclusion}
We introduced GRBO, an RL-based post-training framework, and Warm-K, a test-time sampling strategy, for generative behavior models in autonomous driving. GRBO leverages self-simulation and group-relative rollouts to enhance safety performance while preserving pre-trained human-likeness, whereas Warm-K strategy improves closed-loop execution by aligning motion rollouts for greater behavioral consistency and efficiency without additional training. 

\noindent \textbf{Discussion.}
As shown in the long-tail case (Fig. \ref{fig:demo-long-tail}B), human data are not always expert due to mislabeled samples and imperfect maneuvers, which can hinder IL methods. Hence, we believe RL-based, label-agnostic post-training approaches that preserve pretrained capabilities, like ours, are essential to achieve safer and superhuman performance, one of the fundamental goals of autonomous driving.
Building on our methods, extending Sim Agent models toward generative ego-motion planning with self-policy improvement represents a promising direction for future research.

\noindent \textbf{Limitations.}
Although the effect was marginal, our safety-focused reward design slightly reduced human-likeness during RL post-training. Future work could address this by incorporating realism metrics and broader objectives such as comfort and social compliance. Warm-K also introduces a tunable warm-start parameter $T_w$, which could be further optimized through scenario-specific adaptation.

{
    \small
    \bibliographystyle{ieeenat_fullname}
    \bibliography{main}
}

\clearpage
\setcounter{page}{1}
\maketitlesupplementary
\appendix
\balance 

\section{Supplementary Videos}
We include several demo videos in the supplementary material, each carefully edited to provide additional qualitative support for our method.
\begin{itemize}
    \item \texttt{demo\_1\_RightTurn\_GRBO-1.mp4}: Demonstration of GRBO in the unprotected right-turn scenario (Fig. 4(A)). Our model generated conservative maneuvers to avoid collisions with both vehicles and cyclists. 
    
    \item \texttt{demo\_1\_RightTurn\_GRBO-2.mp4}: Another GRBO demonstration in the same unprotected right-turn scenario (Fig. 4(A)). Our model performs stop-and-go maneuvers to avoid collisions with both vehicles and cyclists.
    
    \item \texttt{demo\_1\_RightTurn\_SMART.mp4}: Demonstration of SMART in the unprotected right-turn scenario (Fig. 4(A)). This pure IL-based method fails to account for interactions with dense traffic, resulting in severe collisions.

    \item \texttt{demo\_2\_WrongWay\_GRBO-1.mp4}: Demonstration of GRBO in the long-tail wrong-way driving scenario (Fig. 4(B)). The lower agent successfully avoided a collision with the oncoming vehicle and recovered to a valid branch of the intersection.

    \item \texttt{demo\_2\_WrongWay\_GRBO-2.mp4}: Another GRBO demonstration in the same long-tail wrong-way driving scenario (Fig. 4(B)). The upper agent avoids a collision with the lower vehicle by performing a backward-driving maneuver, which is an emergent behavior arising from RL-based post-training.

    \item \texttt{demo\_2\_WrongWay\_SMART-1.mp4}: Demonstration of SMART in the long-tail wrong-way driving scenario (Fig. 4(B)). This pure IL-based method attempted to produce collision-avoidance maneuvers but ultimately resulted in tight maneuvers that led to collisions.

    \item \texttt{demo\_2\_WrongWay\_SMART-2.mp4}: Another SMART demonstration in the same long-tail wrong-way driving scenario (Fig. 4(B)). The upper agent failed to decelerate, resulting in a collision with the lower vehicle.

    \item \texttt{demo\_closed\_Straight\_TopK.mp4}: Demonstration of the pure Top-K sampling method in the straight-driving scenario of the closed-loop evaluation (Fig. 5(A)).

    \item \texttt{demo\_closed\_Straight\_WarmK.mp4}: Demonstration of the Warm-K hybrid sampling method in the same straight-driving scenario of the closed-loop evaluation (Fig. 5(A)).

    \item \texttt{demo\_closed\_TurnRight\_TopK.mp4}: Demonstration of the pure Top-K sampling method in the right-turn scenario of the closed-loop evaluation (Fig. 5(C)).

    \item \texttt{demo\_closed\_TurnRight\_WarmK.mp4}: Demonstration of the Warm-K hybrid sampling method in the right-turn scenario of the closed-loop evaluation (Fig. 5(C)).
    
\end{itemize}

\section{Implementation Details}
We trained and evaluated our method on the Waymo Open Motion Dataset (WOMD) \citep{ettinger2021large}. Following SMART \citep{wu2024smart}, we adopted its architecture as our policy model and pre-trained it for 32 epochs on the full dataset. For post-training, we fine-tuned the model with RL using only 10\% of the data for 10 epochs. Gradient accumulation was applied during group-sampling RL to match the compute budget. Thus, the effective batch size was 80 on 8 × A100 GPUs, matching the configuration used in the SMART and CAT-K \citep{zhang2025closed} baseline methods. Closed-loop evaluations were conducted in Waymax \citep{gulino2023waymax}, a multi-agent interactive simulator.
Table~\ref{tab:hyperparam} summarizes the hyperparameter settings used in our GRBO-based post-training.

\begin{table}[h]
\centering
\begin{tabular}{lc}
\toprule
Parameter & Value \\
\midrule

    Batch size& 80\\
    Epoch& 10\\
    Number of rollouts $G$& 8\\
    KL weight $\beta$& 0.1\\
    clip-low $\epsilon_l$& 0.2\\
    clip-high $\epsilon_h$& 0.4\\
    Warm-k sampling steps $T_w$& 2\\

\bottomrule
\end{tabular}
\caption{Hyperparameter configuration.}
\label{tab:hyperparam}
\end{table}


\end{document}